# Physiology-Aware CNN and Zero-Shot Multimodal LLMs for ECG Image Classification: A Comparative Study

Khalil Ahammad [a, b, c], Derek Abbott [d], and Mohsen Dorraki [a, b, c]

[a] School of Computer Science and Information Technology, Adelaide University, SA 5005, Australia
[b] Australian Institute for Machine Learning (AIML), Adelaide, Australia
[c] Pi MedTech, Adelaide, Australia
[d] School of Electrical and Electronic Engineering, Adelaide University, SA 5005, Australia

## Abstract

Multimodal large language models (LLMs) are increasingly adopted to interpret 12-lead ECG images, though the interpretations often lack validation. However, ECG image understanding significantly differs from general images as it depends on precise waveform morphology, lead relationships and accurate interval measurements. This study investigated whether zero-shot multimodal LLMs can reliably distinguish normal and abnormal ECG images and, in parallel, evaluated CNN-based models for clinically grounded references. Standard 12-lead ECG recordings were rendered as single-page images for a binary normal-abnormal classification task. Three prominent LLMs (GPT-5.2, GPT-4.1, and Gemini-2.5 Pro) were tested using a fixed zero-shot prompt across multiple runs. In parallel, a physiology-aware CNN-based model was developed with the capability to aggregate features from the predefined anatomical lead groups. The model was compared with ResNet18, DenseNet121, VGG16 baselines, and all the models were evaluated on an internal test set and external PTB-XL dataset. Across seeds, CNN-based models demonstrated stable discrimination, with average internal ROC-AUC of 0.92–0.94, and external ROC-AUC of 0.85–0.86. The proposed LeadGroupECG model significantly improved over its backbone internally without compromising external generalization. It remained competitive with other baselines, while consistently highlighting anatomical lead-group contributions. In contrast, zero-shot LLM discrimination remained near-chance (ROC-AUC around 0.5). The PR-AUC improved slightly when ECGs used a grid-based calibration background compared with the grid-free ECGs. Although multimodal LLMs can generate reasonable ECG narratives, their zero-shot diagnostic discrimination remains limited. Therefore, clinically framed, domain-specific architectures remain essential for AI-based ECG interpretation.

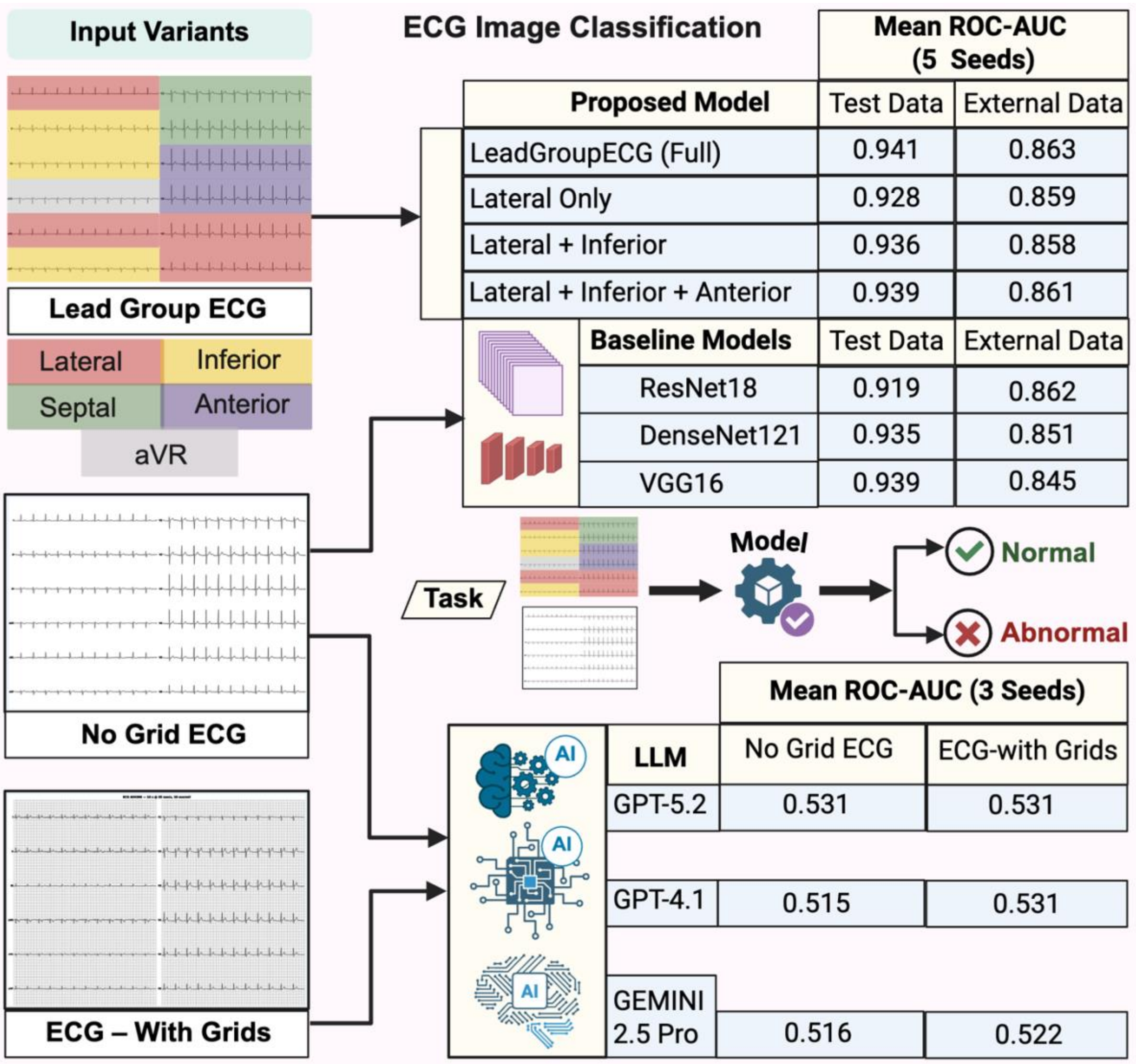

**Graphical Abstract**

## 1. Introduction

Electrocardiogram (ECG) analysis is one of the primary tools widely used in cardiovascular diseases management aiding in rapid assessment of heart rhythm, conduction, and structural abnormalities [1, 2]. In recent years, deep learning has demonstrated substantial potential in ECG abnormality detection, with convolution neural networks (CNNs) sometimes achieving cardiologist-level performance when trained on large-scale raw waveform signal datasets [2–8]. The signal-based approaches usually exploit the precise temporal resolution and amplitude information while extracting the morphological features [9–12]. However, in real-world clinical practice, ECGs are often reviewed or shared as rendered images rather than structured signal files. Consequently, image-based ECG analysis has emerged with great significance for AI-based ECG interpretation.

Note that ECG images vary in layouts, grid calibration, resolution and visual artifacts. Thus, image-based models need to contend with these image variations because they can influence morphological interpretation and downstream analysis [13, 14]. Several studies have applied deep learning architectures to ECG images with encouraging outcomes [15–18], and they treat the ECG sheet as a generic visual object without explicit consideration of the anatomical relationships among the leads. However, in clinical interpretations, the leads of a 12-lead ECG are anatomically grouped to produce separate viewpoints of the inferior, lateral, septal and anterior walls of the heart [19]. Thus, incorporating these lead groups into a model to perform a coordinated assessment across the viewpoints may enhance the model performance and interpretability.

Parallel to the advancements in deep learning-based ECG image analysis, multimodal LLMs have gained rapid attention in ECG image processing because of their capability in generating diagnostic-style explanations from visual inputs. However, ECG image analysis is fundamentally different from natural image processing. Semantic visual recognition alone is not sufficient for accurate ECG image interpretation; rather, it requires precise spatial morphology analysis, interval measurements, axis assessment and evaluation of the inter-lead relationships [20]. Emerging evidence suggests that the performance of multimodal LLMs in specialized diagnostic tasks may remain limited without task-specific training [21]. For instance, an evaluation of the GPT-4o model on a reasonably small ECG image dataset demonstrated that the accuracy was significantly enhanced when evaluation was shifted from zero-shot to few-shot learning [22]. However, positioning LLMs as structured assistants for generating contextual descriptions, retrieving domain knowledge, and guiding representation learning rather than implementing as standalone diagnostic models for ECG interpretation was found effective in several studies [23–27]. They implemented LLMs either as an interpretable reasoning layer within a broader ECG analysis pipeline or as a validation layer to support clinically grounded interpretations.

The rapid expansion of multimodal LLM capabilities has generated increasing interest in their use for ECG interpretation. In this study, we aim to evaluate the state-of-the-art multimodal models on a comparatively large ECG image dataset for a binary classification task as a foundational benchmark. The outcomes of this investigation provide significant insights into whether the current LLMs are sufficiently reliable for a complete ECG interpretation that requires precise diagnosis categorization and subclass identification of the abnormalities. The objectives of this study can be summarised as follows: (a) to establish baseline performance of the traditional CNN-based models for the binary classification of ECG images; (b) to evaluate a physiology-aware CNN-based model (LeadGroupECG) designed to capture the anatomical relationships among the lead groups; and (c) to assess the state-of-the-art multimodal LLMs for the same task under standardized zero-shot conditions. The graphical abstract also depicts a summary of the approach and the key findings of the study.

## 2. Materials and methods

The overall study framework is designed to evaluate ECG images using both domain-specific CNN-based models and generalist multimodal LLMs. This section consists of three major components: data preparation and standard 12-lead image rendering, supervised classification using baseline models along with the proposed LeadGroupECG model, and zero-shot evaluation of the LLMs. All experiments were conducted using multiple independent seeds and identical image inputs. LeadGroupECG and the baselines models were trained and evaluated on google colaboratory [28] using identical training and evaluation protocol. The graphical abstract and figures 1–4 were created at BioRender.com.

### 2.1 Data collection and preparation

#### 2.1.1 Data sources

We selected a large-scale public dataset [29] for this study. The original dataset is acquired from Shandong Provincial Hospital that consists of 25,770 ECG records (12-lead) with varying lengths from 10 to 60 seconds. The dataset provides 44 standardized primary diagnostic statements, and we consider statement ‘Normal ECG’ (Code = 1) as the ‘Normal’ class and the rest of the statements together as ‘Abnormal’ class. We denoted this dataset as ‘SPH-data’ in this article. In addition, we used PTB-XL, another publicly available dataset [30] for external validation purpose. Note that PTB-XL precisely defines five super classes of diagnosis and 23 standardized diagnostic statements. Again, we define the normal and abnormal classes based on the provided codes in the database, and we denoted this dataset as ‘PTB-XL’.

#### 2.1.2 Data preparation

The SPH-data was filtered to keep the records of length 10 seconds only for a stable image rendering experience and to match with PTB-XL, because PTB-XL provides recordings of 10 seconds length only. All the duplicates were removed to protect any potential data leakage. The normal class was slightly down-sampled randomly (stratified by age group) around the abnormal class to increase balance and reduce evaluation cost of the LLM models. Only the test fold (Fold = 10) of PTB-XL was used for external validation, where only the diagnoses with 100% physician validation of illness findings (similarly as [31]) were considered. The finalized SPH-data was randomly split for training, internal validation, and testing the models. We used 80% of the data for training, 10% for internal validation and preserved the remaining 10% for testing. The final dataset size and the overall distributions are summarized in Table 1.

**Table 1.** Summary of dataset composition and class distribution across training, validation, internal testing and external validation cohort. **N** is for Normal and **AB** represents the Abnormal class in the class column under sample distribution.

| **Datasets** | **Total** | **Normal ECGs** | **Abnormal ECGs** | **Sample Distribution** | | | |
|---|---|---|---|---|---|---|---|
| | | | | **Class** | **Train** (80%) | **Validation** (10%) | **Test** (10%) |
| SPH-data | 14871 | 7500 | 7371 | **N** | 6000 | 750 | 750 |
| | | | | **AB** | 5896 | 737 | 738 |
| PTB-XL (Test Fold) | 1685 | 615 | 1070 | **N** | – | – | 615 |
| | | | | **AB** | | | 1070 |

#### 2.1.3 ECG image rendering

Here, 12-lead ECG recordings were converted into standardized single page images using a fixed clinical layout. The signals were plotted in a $6 \times 2$ lead layout (limb leads in the left column and precordial leads in the right) at a sampling frequency of 500 Hz over 10 seconds duration using fixed clinical calibration parameters (paper speed: 25 mm/s, and gain: 10 mm/mV). Rendering was performed in millimetre-scaled coordinates to maintain consistent spatial morphology and inter-lead relationships. Images were initially generated at high resolution and down-sampled to a fixed dimension ($908 \times 1280$) before training and evaluation. Two variants were generated: a grid-based clinical format and a grid-free version. Fig. 1 demonstrates two sample rendered images.

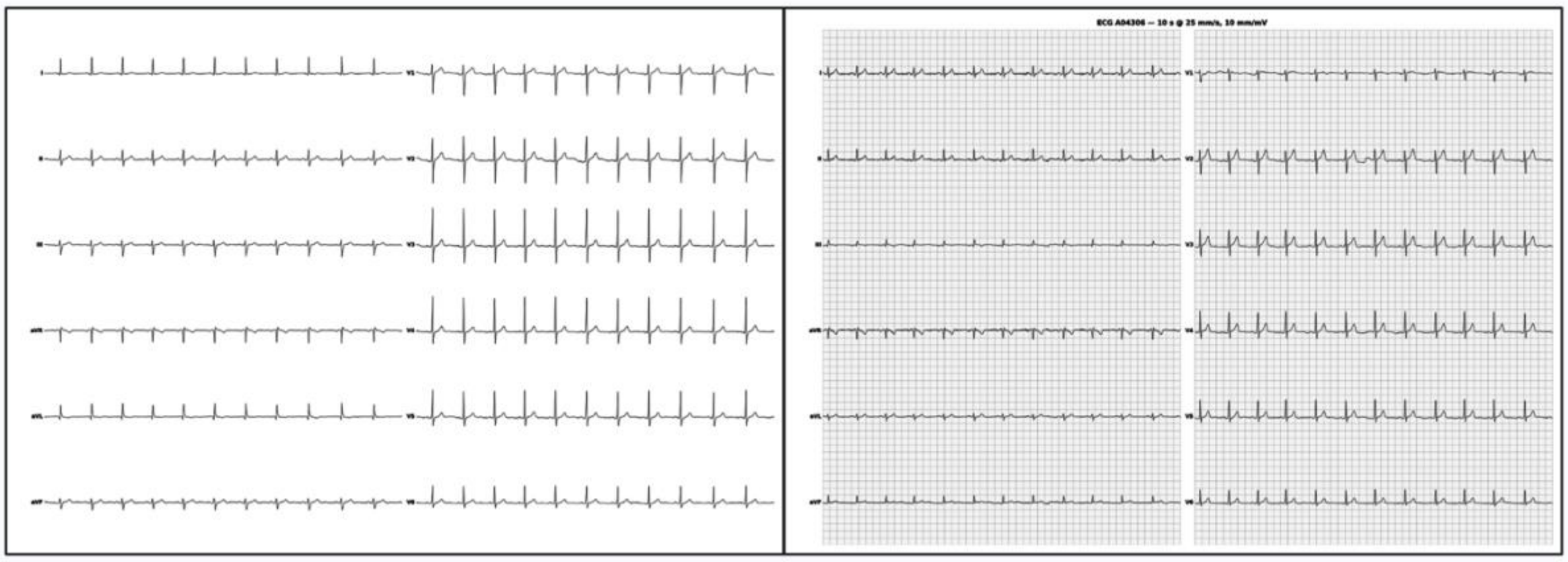

**Fig. 1.** Sample rendered ECG images with grid-free and grid-based backgrounds are shown here. Each image consists of two columns: lead I, II, III, aVR, aVL, and aVF in the left column, and V1-V6 in the right.

## 2.2 CNN-based models

### 2.2.1 Baselines architectures

To establish reference performance on the same datasets for this classification task, three widely used CNN-based architectures were evaluated: ResNet18 [32], DenseNet121 [33], and VGG16 [34]. These models were selected because of their varying architectural inductive biases, such as, residual learning, dense connectivity, and sequential convolutional design. Also, these architectures were used in different ECG associated tasks to achieve improved performances [35–38]. Each of these backbones was adapted for single-channel grayscale ECG inputs. These baseline models were implemented to treat the ECG page as a visual object with a view to exploiting the global morphologies of an ECG image. The implementation pipeline consists of uniform preprocessing, optimization schedules, and evaluation criteria.

### 2.2.2 Proposed model: LeadGroupECG

LeadGroupECG is a physiology-aware convolutional architecture for ECG image classification that embeds anatomical lead-group structure directly into the representation learning process. Instead of treating the ECG page as a single homogeneous image, the model decomposes the spatial layout into clinically structured regions and integrates them through a dual-pathway decision framework. This design enables to identify the region that provides the strongest signal and quantify the sickness of each wall of the heart. In this design, we preserve the physiological relationships among the leads while maintaining compatibility with conventional convolutional optimisation. The general overview of the model is illustrated in Fig. 2.

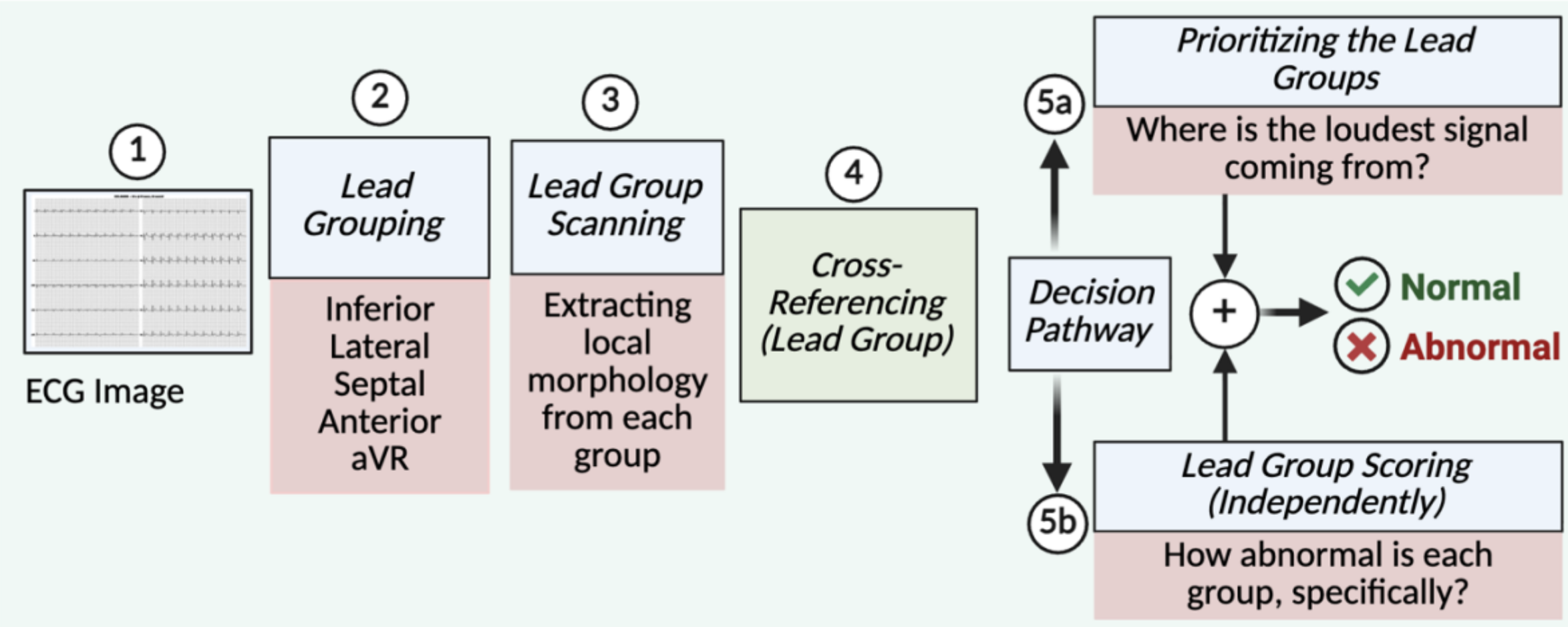

**Fig. 2.** LeadGroupECG treats the ECG image as a group of clinically structured views where each view represents a specific wall of the heart.

We denote a batch of grayscale ECG images as $X \in \mathbb{R}^{B \times 1 \times H \times W}$, where $B$ is the batch size and each ECG image has a fixed dimension of $H \times W$. Given the standardized $6 \times 2$ ECG layout, a deterministic lead-group extractor is applied to isolate anatomically meaningful lead regions and reorganize them into five parallel lead group views corresponding to the inferior, lateral, septal, anterior, and aVR territories, as illustrated in Fig. 3. Each extracted lead-group is padded to a common spatial resolution $(H_t, W_t)$. The extractor $E$ maps the input ECG batch into anatomical lead-group views and an associated validity mask as follows:

$$(V, M) = E(X), \qquad \text{where } V \in \mathbb{R}^{B \times 5 \times 1 \times H_t \times W_t}, M \in \{0,1\}^{B \times 5} \tag{1}$$

Here, $V$ denotes the extracted lead-group tensor, whereas the binary mask $M$ indicates which lead groups are available for each ECG sample, enabling the model to handle incomplete or missing regions without disrupting downstream processing.

Each anatomical lead-group view is then encoded by a shared Resnet18 backbone with an output stride of 32. Before convolutional encoding, the lead-group dimension is flattened into the batch dimension, resulting in $V' \in \mathbb{R}^{(B.5) \times 1 \times H_t \times W_t}$. The shared encoder $\phi(\cdot)$ subsequently produces convolutional feature representations according to Eq. (2)

$$F = \phi(V') \in \mathbb{R}^{(B.5) \times 512 \times H_t/32 \times W_t/32}. \tag{2}$$

To refine these activations, an identity-safe residual convolutional block attention module (CBAM) is incorporated without substantially altering the pretrained backbone feature statistics. The refined feature maps are then passed through adaptive average pooling with a $1 \times 1$ spatial output to generate compact descriptors. These descriptors are linearly projected and layer normalized to form lead-group tokens $T \in \mathbb{R}^{B \times 5 \times d}$, where $d = 256$. A learnable group embedding is further added to preserve the lead-group identity after reshaping.

The model allows controlled interactions among the lead groups. The tokens are passed through a single mask-aware interaction block consisting of masked multi-head-self-attention (MHSA) followed by a feed-forward multilayer perceptron with residual connections. The validity mask $M$ from Eq. (1) is propagated throughout this stage to ensure that unavailable lead groups do not contribute to token interactions. This process produces a refined token representation, denoted as $T'$.

Predictions are generated through two complementary pathways operating on $T'$. In the adaptive pooling pathway, token-wise scalar scores are produced by a lightweight two-layer feed-forward network and converted into temperature-scaled weights using softmax normalization. The resulting weights are used to aggregate the lead-group tokens into a global ECG representation using Eq. (3)

$$z = \sum_{i=1}^{5} w_i T_i'. \tag{3}$$

During training, a bounded stochastic scaling is applied to $z$ to improve the robustness of the model. In parallel, a concept bottleneck model (CBM) pathway maps each refined lead-group token to a scalar concept score through linear projection. These concept scores represent region-specific abnormality evidence and are modulated by a learnable positive scaling factor. The lead-group concept vector is computed according to Eq. (4), where $g(\cdot)$ denotes the linear concept projection and $\beta$ is a learnable positive scaling parameter. This pathway provides an anatomically interpretable representation of abnormality contributions across the five lead-group territories

$$c = \beta g(T') \in \mathbb{R}^{B\times 5}, \beta > 0. \tag{4}$$

Finally, the pooled representation $z$, and the scaled concept scores $c$ are concatenated to construct the final decision representation as follows:

$$y = h(f) \in \mathbb{R}^{B\times 2}, \qquad where\ f = [z; c] \in \mathbb{R}^{B\times(d+5)}. \tag{5}$$

Here, $h(\cdot)$ denotes the final linear classifier that produces the binary logits for normal-abnormal ECG classification. The complete workflow of this model is summarized in Fig. 4.

### 2.2.3 Training and evaluation protocol

All CNN-based models were trained within a unified optimisation framework to ensure a justified architectural comparison. The same data preprocessing, optimisation strategy, model-selection criterion, and evaluation protocol were applied across the baseline CNN architectures and the proposed LeadGroupECG model.

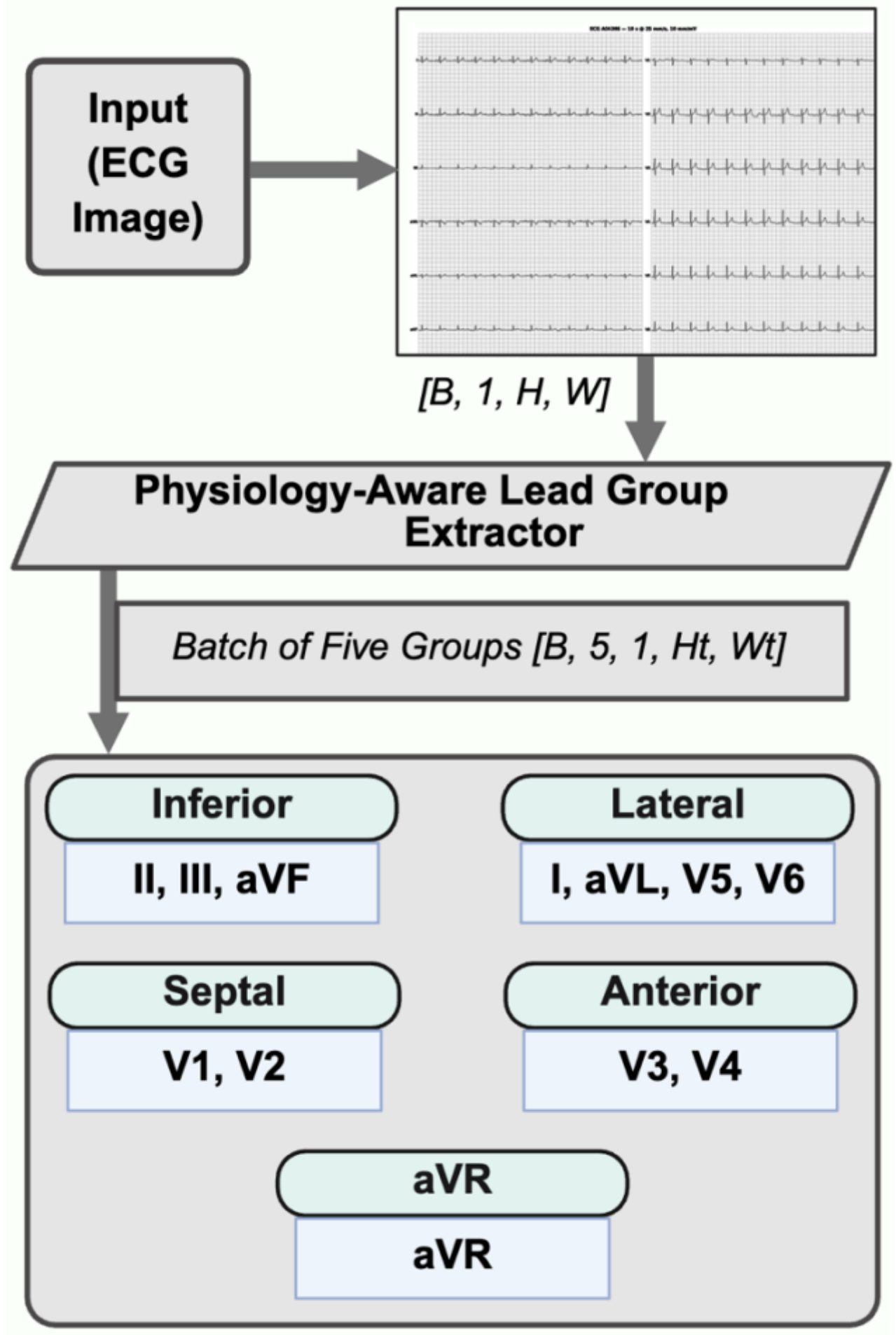

**Fig. 3.** Physiology-aware lead group extractor creates a batch of five anatomical lead groups from an ECG image.

Model parameters were optimised using categorical cross-entropy loss for binary normal–abnormal ECG image classification, with abnormal ECGs defined as the positive class. For a given sample, the loss function was defined using Eq. (6), where $p_c$ denotes the predicted probability for class $c$, and $y_c$ represents the corresponding one-hot encoded ground-truth label

$$L_{\mathrm{CE}} = -\sum_{c=1}^{2} y_c \log(p_c). \tag{6}$$

Optimisation was performed using AdamW with decoupled weight decay. To stabilise transfer learning from ImageNet [39] initialisation, backbone parameters were updated with a lower effective learning rate than newly introduced task-specific layers. The learning-rate schedule consisted of an initial linear warm-up phase followed by cosine decay. Model selection was based on the validation loss rather than test-set performance. Specifically, an exponential moving average (EMA) of the validation loss was monitored to reduce sensitivity to epoch-level fluctuations. Early stopping was triggered when no improvement in the EMA-smoothed validation loss was observed within a predefined patience window. The model checkpoint with the best validation performance was retained for final evaluation.

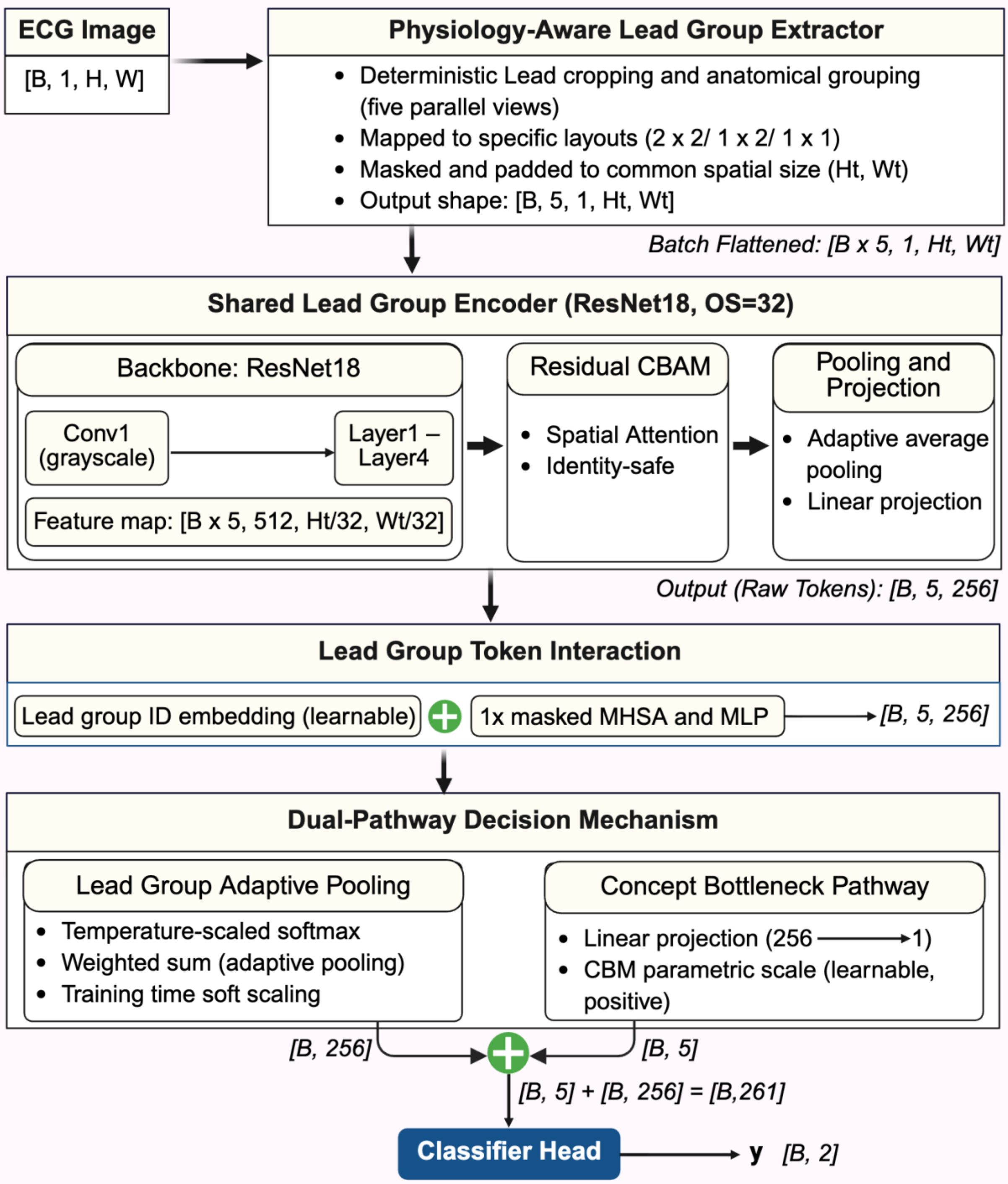

**Fig. 4.** The complete workflow of LeadGroupECG is illustrated in this figure that consists of an anatomical lead-group extractor, a shared encoder, a lead-group token mixer, and a concatenated decision pathway of lead-group adaptive pooling and region specific CBM scores.

Final evaluation was conducted on the unseen internal test set of the SPH-data and on the external PTB-XL test-fold cohort without further optimisation, retraining, and threshold tuning. Class probabilities were obtained using the softmax output of each model. Discriminative performance was quantified using threshold-independent metrics, including ROC-AUC and PR-AUC. In addition, threshold-based classification performance was assessed using precision, recall, and F1-score at a fixed probability threshold of 0.5. All experiments were repeated across fixed independent random seeds, and results are reported as mean ± standard deviation to reflect both average performance and run-to-run variability.

### 2.3 LLM models

To evaluate the capability of multimodal LLMs in ECG abnormality detection, we implemented a standardized zero-shot image interpretation approach for three state-of-the-art vision-language models: GPT-5.2, GPT-4.1, and Gemini 2.5 Pro. The objective was to assess whether general-purpose vision language models could identify abnormal electrocardiographic patterns from rendered ECG images without task-specific training, fine-tuning, or few-shot examples. The core focus was on evaluating the intrinsic multimodal reasoning capability of the models rather than optimizing them.

The LLM evaluation was conducted on the test cohort of internal SPH-data, comprising 1,488 ECG images, including 750 normal and 738 abnormal ECGs. Each ECG was evaluated under two image-rendering conditions. In the grid-free setting, waveform traces were displayed without a calibration grid. In the grid-based setting, ECGs were rendered with clinically standard millimetre-scale calibration grids. The same internal test cohort was assessed separately under both rendering protocols.

A fixed zero-shot prompt was used across all models, rendering conditions, and repeated runs to ensure experimental consistency. The prompt instructed each model to classify the ECG as normal or abnormal and provide a confidence score between 0 and 100. Strict JSON-only responses were enforced to reduce ambiguity in output parsing. Each ECG image was submitted individually to the corresponding multimodal LLM through the vendor-provided API interface, corresponding to an inference batch size of one image per request. The image transmission format differed across providers: OpenAI models received base64-encoded image data URLs, whereas Gemini received raw image bytes. The core experimental configuration, including the evaluation cohort, rendering conditions, prompt, image transmission format, inference batch, and output structure is summarized in Table 2.

Three independent evaluation runs were performed using fixed random seeds while preserving the same test cohort and prompt structure. Model outputs were converted into probabilistic scores by mapping confidence values to class probabilities. Performance was evaluated using both threshold-independent metrics (ROC-AUC and PR- AUC), and threshold classification metrics (precision, recall and F1-score) at a fixed probability threshold 0.5.

**Table 2.** Summary of the zero-shot multimodal LLM evaluation protocol.

| Protocol | Specification |
|---|---|
| **Evaluation Cohort** | Test set of SPH-Dataset: 1,488 ECG images, including 750 normal and 738 abnormal ECGs. |
| **Rendering Conditions** | Grid-free and grid-based 12-lead ECG images evaluated separately. |
| **Models** | GPT-5.2, GPT-4.1, and Gemini-2.5 Pro |
| **Evaluation Setting** | Zero-shot ECG image interpretation; no few-shot examples, fine-tuning, or prompt optimization. |
| **Fixed Prompt** | "You are interpreting a standard 12-lead ECG image. Classify this ECG image as Normal or Abnormal. Provide a confidence score from 0 to 100. Return only JSON in the format: {"label":"Normal","confidence": }<br>Or {"label":"Abnormal","confidence": }." |
| **Image Transmission Format** | OpenAI: base64-encoded image data URLs, Gemini: raw image bytes. |
| **Inference Batch** | One ECG image per API request. |
| **Repeated Runs** | Three independent seeded runs using the same test cohort. |

### 3. Results

Across multiple independent runs and evaluation on both internal and external datasets, several key patterns were observed. First, conventional CNN-based models achieved strong and consistent performance for this classification task, with only a modest performance decay when tested on the external PTB-XL dataset. Second, the anatomically structured LeadGroupECG model provided a statistically significant improvement over its backbone architecture (ResNet18) on internal test, while maintaining comparable performance on PTB-XL and other baseline models. Third, zero-shot multimodal LLMs performed closed to chance-level, irrespective of two specific rendering styles. These results suggest that domain-specific structural models are more effective than general-purpose multimodal reasoning.

### 3.1 Evaluation of CNN-based models

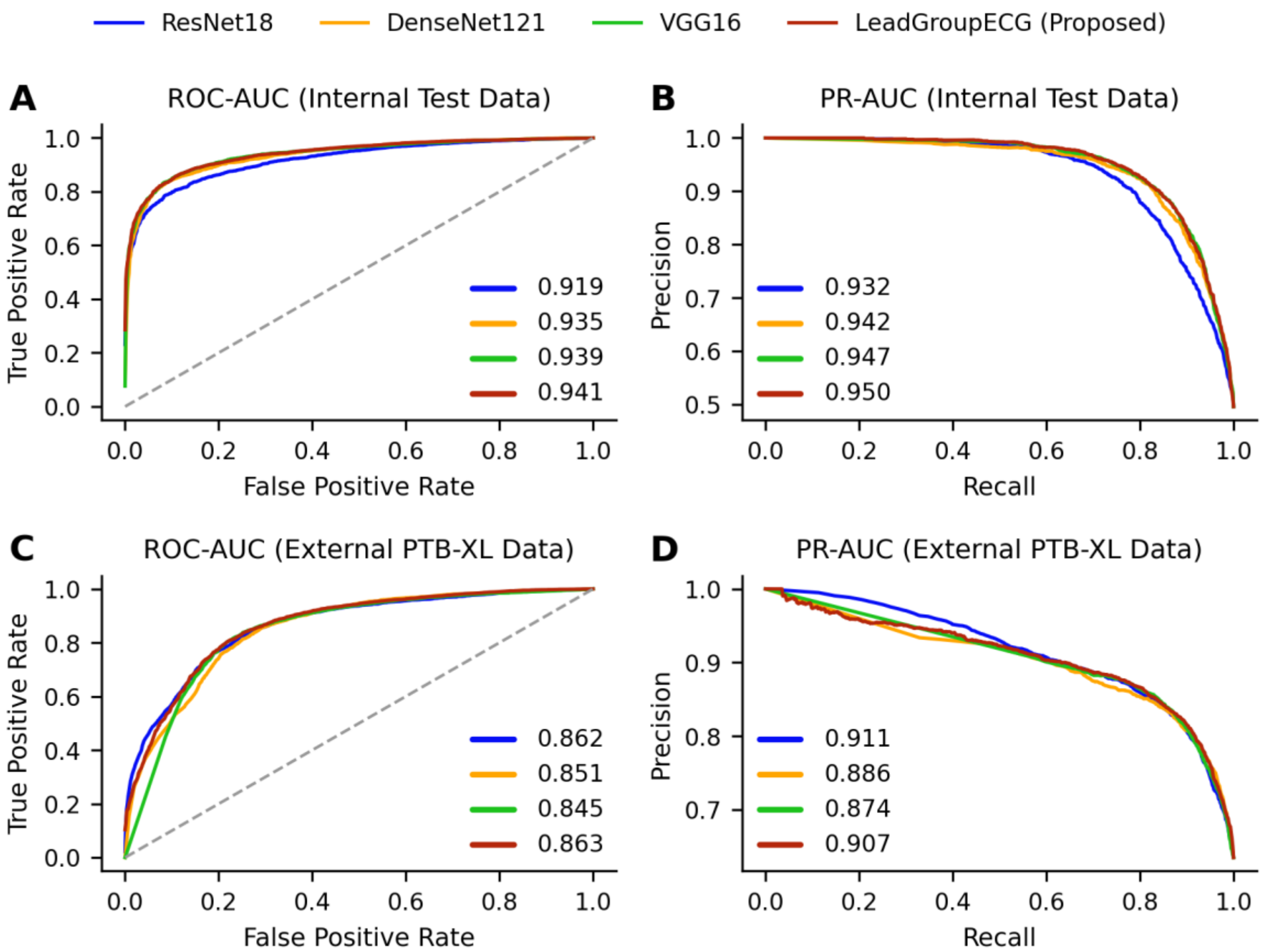

**Fig. 5**. The performance of the CNN-based models including LeadGroupECG is exhibited here in terms of ROC-AUC and PR-AUC on both internal data (A and B), and external

On the internal test cohort, all the baseline CNNs demonstrated high and closely clustered discrimination (minimum ROC-AUC ≈ 0.92, and PR-AUC ≈ 0.93). The ROC and precision-recall curves (Fig. 5A, 5B) were tightly concentrated in the high-performance region, and a small variability across seeds indicates stable optimisation. The narrow performance gap between the strongest and weakest model suggests that when ECG sheet images are rendered under a standardized physical pipeline, they are sufficiently structured to allow multiple CNN backbones to reach near-ceiling discrimination. On external PTB-XL validation (Fig. 5C, 5D), all CNNs demonstrated a moderate reduction in ROC-AUC (minimum ≈ 0.85), reflecting the cross-dataset distributional differences, and the impact of the slightly imbalanced class scenarios. No single architecture showed disproportionate degradation, supporting the robustness of the rendering strategy and training protocol. Table 3 presents the overall performance of all the CNN-based models.

### 3.2 Evaluation of LeadGroupECG

This proposed model achieved the highest internal discrimination among the evaluated CNN models. The internal mean ROC-AUC reached 0.941 ± 0.002 that is significantly better than its backbone architecture (ResNet18). The average improvement in ROC-AUC across seed was +0.022 (95% CI 0.0129 – 0.0304, paired $p$ = 0.0023), indicating a statistically reliable achievement. The gain was accompanied by increased sensitivity and specificity, suggesting a better decision boundary toward improved detection of abnormal cases. At the same time, the generalization of LeadGroupECG on PTB-XL was preserved as the discrimination was found nearly identical with ResNet18.

Analysis of the lead-group adaptive pooling weights (Fig. 6) revealed a stable hierarchy of territorial contribution. Across seeds, the lateral lead-group consistently received the highest mean weight, followed by the inferior and anterior territories, whereas septal and aVR contributed less prominently. Although the lateral view frequently provided the strongest influence, non-zero weights across other territories demonstrated distributed integration. The model appeared to the lead groups that contain most informative features, while retaining complementary signals from the other lead groups. This adaptive mechanism therefore functioned as a structured aggregation strategy that remained anatomically coherent.

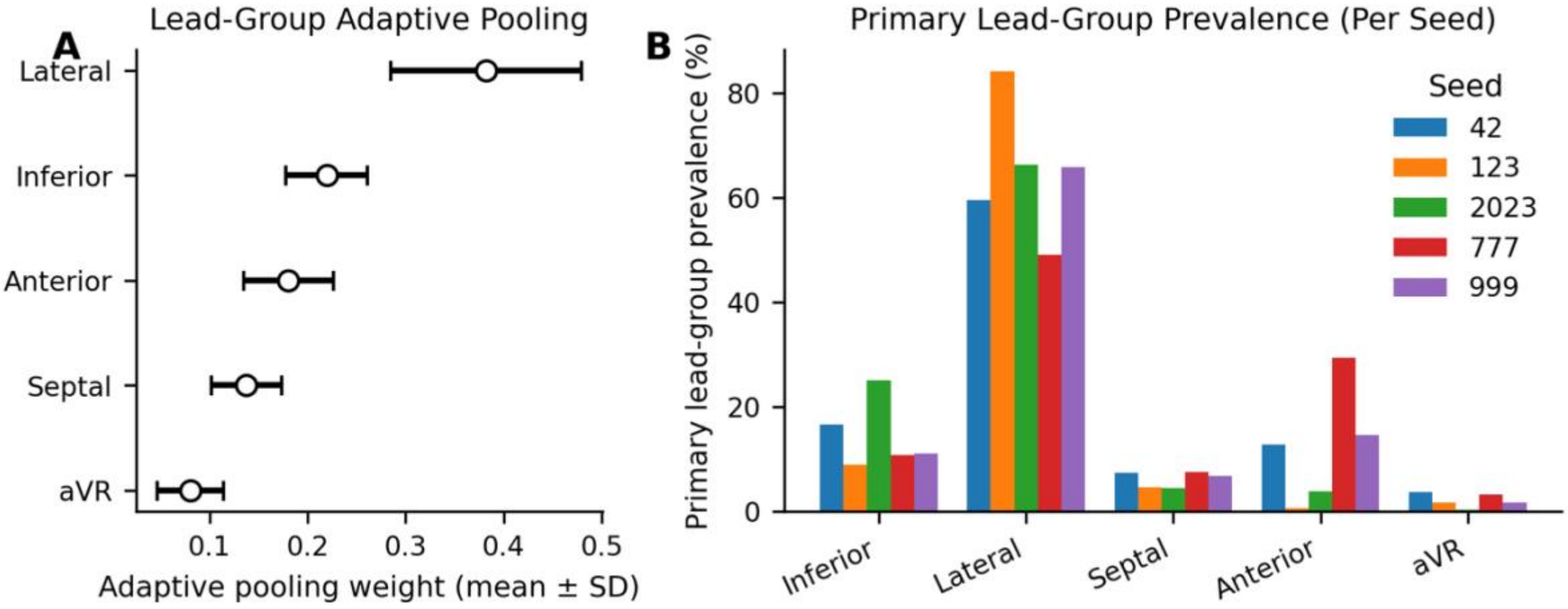

**Fig. 6.** Adaptive lead-group weighting and territorial dominance in LeadGroupECG. **(A)** Mean adaptive pooling weights (± SD) assigned to each anatomical lead group across five seeds. (**B**) Percentage of test samples in which each lead group was the highest-weighted contributor per seed.

**Table 2**: The overall performance of the CNN-based models using both threshold-independent (ROC-AUC, PR-AUC) and threshold-based metrics (Precision, Recall, F1-Score) at a fixed threshold of 0.5 on both internal and external data is listed in this table. All the metrics are expressed as mean ± standard deviation of five independent runs and evaluations.

| Internal Test (SPH-Dataset) | | | | | |
|---|---|---|---|---|---|
| **Model** | **Precision** | **Recall** | **F1-Score** | **ROC-AUC** | **PR-AUC** |
| ResNet18 | 0.833±0.080 | 0.831±0.079 | 0.826±0.003 | 0.919±0.006 | 0.932±0.005 |
| DenseNet121 | 0.888±0.037 | 0.836±0.043 | 0.860±0.012 | 0.935±0.010 | 0.942±0.009 |
| VGG16 | 0.889±0.029 | 0.852±0.029 | 0.869±0.007 | 0.939±0.007 | 0.947±0.005 |
| LeadGroupECG | 0.853±0.027 | 0.883±0.015 | 0.868±0.007 | 0.941±0.002 | 0.950±0.002 |
| **External Validation (PTB-XL)** | | | | | |
| ResNet18 | 0.821±0.036 | 0.867±0.073 | 0.841±0.018 | 0.862±0.017 | 0.911±0.013 |
| DenseNet121 | 0.826±0.019 | 0.870±0.031 | 0.847±0.010 | 0.851±0.032 | 0.886±0.044 |
| VGG16 | 0.828±0.013 | 0.875±0.025 | 0.850±0.008 | 0.845±0.014 | 0.874±0.015 |
| LeadGroupECG | 0.821±0.014 | 0.891±0.018 | 0.855±0.006 | 0.863±0.015 | 0.907±0.019 |

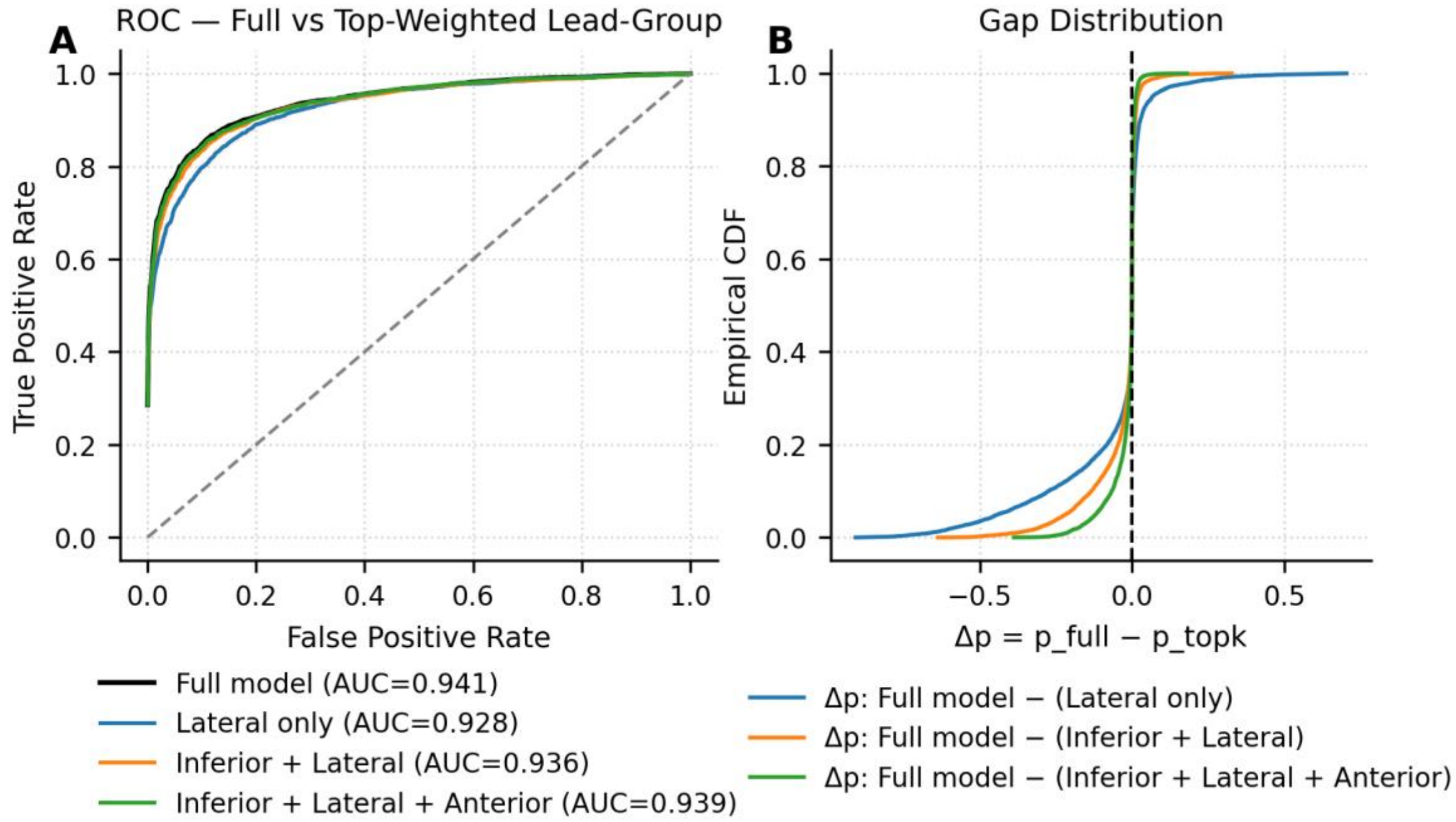

**Fig. 7**. Sufficiency of dominant lead-group vs full model. (**A**) ROC curves comparing the full LeadGroupECG model with progressively restricted models using only the highest weighted lead groups. Modest decrease in performance with territory reduction is observed, and it is restored with incremental reintegration. (**B**) Empirical cumulative distribution of the probability differences (Δp = p_full – p_topk) is depicted. Most samples cluster near zero, indicating substantial signal is captured by dominant views.

The sufficiency testing further clarifies the role of lead-group integration (Fig. 7). When predictions were restricted to the dominant lateral group alone, internal ROC-AUC decreased from 0.941 to 0.928. Probability gap analysis showed that for many samples, restricting inference to the primary territory produced only small shifts in predicted probability, indicating that a dominant region captures a substantial portion of the discriminative signal. However, secondary lead groups also contribute meaningful complementary information.

### 3.3 Evaluation of multimodal LLMs

In contrast to the CNN-based models, multimodal LLMs demonstrated near-chance level discrimination (ROC-AUC ≈ 0.51 – 0.53 across GPT-4.1, GPT-5.2, and Gemini-2.5 Pro) for both grid-free and grid-rendered ECG images. Although GPT-5.2 exhibited higher recall (0.75 on the grid-free images and 0.83 on the grid-based images.). However, class separation was not improved due to low precision. These results suggest that zero-shot multimodal reasoning, even with advanced foundation models, does not capture adequate spatial and morphological features required for ECG image classification. The model capabilities are presented in Fig. 8 using ROC-AUC and PR-AUC. Table 3 provides a detailed view of both threshold-free and threshold-based metrics.

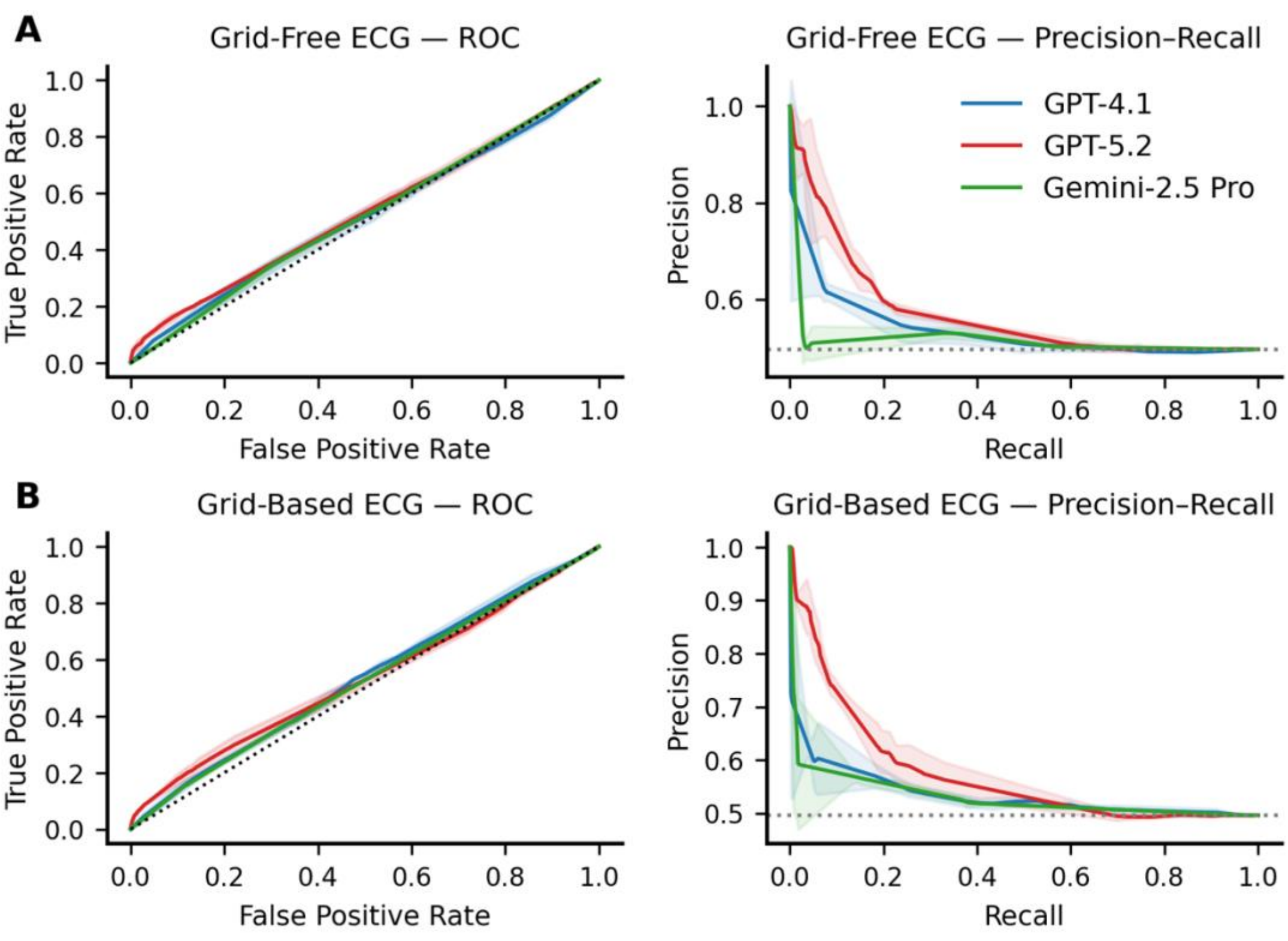

**Fig. 8.** ROC and precision-recall curves on **(A)** grid-free and **(B)** grid-based ECG images. Performance remains close to chance.

**Table 3**: The performance of the multimodal LLMs using both threshold-independent (ROC-AUC, PR-AUC) and threshold-based metrics (Precision, Recall, F1-Score) at a fixed threshold of 0.5 on both grid-free and grid-based ECG images is shown in this table. All the metrics are expressed as mean ± standard deviation of three independent evaluations.

| **Grid-Free ECG Images** | | | | | |
|---|---|---|---|---|---|
| **Model** | **Precision** | **Recall** | **F1-Score** | **ROC-AUC** | **PR-AUC** |
| GPT-4.1 | 0.501±0.010 | 0.592±0.006 | 0.542±0.007 | 0.515±0.011 | 0.518±0.011 |
| GPT-5.2 | 0.497±0.007 | 0.747±0.008 | 0.597±0.008 | 0.531±0.009 | 0.548±0.008 |
| Gemini-2.5 Pro | 0.503±0.003 | 0.550±0.009 | 0.525±0.003 | 0.516±0.003 | 0.508±0.004 |
| **Grid-Based ECG Images** | | | | | |
| GPT-4.1 | 0.521±0.003 | 0.540±0.008 | 0.530±0.004 | 0.531±0.014 | 0.525±0.013 |
| GPT-5.2 | 0.496±0.002 | 0.826±0.001 | 0.620±0.002 | 0.531±0.014 | 0.548±0.011 |
| Gemini-2.5 Pro | 0.518±0.006 | 0.384±0.007 | 0.441±0.006 | 0.522±0.005 | 0.515±0.004 |

## 4. Discussion

In this study, we performed a structured comparison between conventional CNNs, a physiology-aware CNN architecture, and zero-shot LLMs for binary ECG image classification. The findings of this study provide a clear methodological contrast: rendered ECG images contain sufficient spatial and morphological information for supervised CNN-based models to achieve strong discrimination, whereas current general-purpose multimodal LLMs remain limited when used as standalone zero-shot ECG classifiers. More importantly, the results suggest that successful ECG image interpretation depends not only on visual recognition capacity, but also on morphology-aware and lead-structured inductive biases that align with the clinical logic of 12-lead ECG image interpretation.

First, the strong and stable performance of CNN-based models indicates that rendered ECG images preserve clinically meaningful information in a form that can be exploited by supervised visual learning models. ResNet18, DenseNet121, and VGG16 converged to a similar performance range with modest differences despite their architectural diversity. This convergence suggests that, under controlled rendering conditions, ECG image classification is not solely dependent on architectural scale or complexity. Instead, the fixed spatial layout, high-contrast waveform traces, repeated cardiac cycles, and standardized calibration make ECG images particularly compatible with convolutional inductive biases. This interpretation is consistent with previous ECG deep learning studies showing that task-specific neural networks can extract clinically informative representations from ECG signals and images [2–8, 15–18]. The moderate reduction in performance on external PTB-XL validation is also reasonable due to cross-dataset variation in abnormality patterns, acquisition sources, and slightly imbalanced class representation.

Second, embedding anatomical structure directly into the representation learning process introduced clinically meaningful inductive bias and measurable benefits. LeadGroupECG demonstrated a statistically significant enhancement over its Resnet18 backbone without degrading the external generalizability. This improvement is important because conventional CNNs treat the ECG sheet as a homogeneous visual object, whereas clinical ECG interpretation is dependent on the coordinated evaluation of the changes across anatomically related lead groups representing inferior, lateral, septal, and anterior viewpoints of the heart [20]. By decomposing the ECG page into predefined anatomical lead groups, LeadGroupECG encourages the model to organize discriminative evidence according to clinically meaningful regions. The analysis on the adaptive lead-group weights provides further mechanistic insights into this behaviour. The lateral territory consistently emerged as the dominant contributor across seeds, followed by the inferior and anterior groups. The weighting hierarchy was found stable rather than random suggesting that the model learned physiologically coherent prioritization rather than random feature amplification. At the same time, non-zero weights across the lead groups indicate distributed integration rather than exclusive reliance on specific regions. The sufficiency analysis supports this interpretation: restricting inference to the highest-weighted lateral group produced only a modest reduction in discrimination, while progressive reintegration of additional regions restored performance toward the full model. Therefore, the proposed architecture appears to use a multi-territorial decision strategy in which a dominant lead group captures substantial signal, but secondary territories provide complementary evidence.

Third, in contrast to the CNN-based models, advanced foundational LLMs failed to achieve meaningful discriminatory capacity beyond random assumption across both grid-free and grid-based rendering conditions. The near-chance performance of zero-shot multimodal LLMs highlights a fundamental limitation of general-purpose visual-language reasoning for ECG image classification. Note that ECG interpretation is not a typical semantic image-recognition task. It relies on precise spatial morphology, interval measurements, and coordinated inter-lead analysis [20]. The findings of this investigation are broadly consistent with emerging evidence that multimodal LLM performance in specialized diagnostic tasks remains limited without task-specific adaptation [21, 22]. Prior work has suggested that LLMs may be more effective when integrated into structured ECG pipelines, for example as reasoning layers, report-generation tools, or interpretable interfaces built on top of ECG-specific representations [23–27].

## 5. Limitation of the study

There are several limitations of this study. The classification task was binary, whereas real-world ECG interpretation requires multi-label diagnosis, rhythm analysis, identification of abnormalities, and assessment of clinically subtle findings. The ECG images were generated through a controlled rendering process from structured signals, which may not fully reflect the noise, compression artifacts, scanning distortions, and layout variability encountered in routine clinical ECGs. Although PTB-XL provided external validation, additional validation across multiple institutions, devices, and image acquisition workflows is necessary. This study intentionally evaluated zero-shot multimodal LLMs without few-shot prompting, fine-tuning, or ECG-specific feature extraction. Therefore, the results should not be interpreted as evidence against all LLM-based ECG systems, specially the fine-tuned, tool-augmented, or ECG-specialized LLMs.

## 6. Conclusion

This study demonstrates that ECG image classification benefits from domain-specific structural design rather than generic multimodal reasoning. Conventional CNN architectures can achieve strong and reproducible discrimination when trained on standardized ECG renderings. Incorporating anatomical lead-group structure through the proposed LeadGroupECG architecture provides a statistically significant model that integrates multi-territorial signals in a physiologically coherent manner. On the other hand, despite advanced visual language capabilities, zero-shot LLMs exhibit minimal discrimination, consistent with near-random performance. These results indicate that ECG classification requires morphology-aware inductive bias and cannot be reliably achieved through multimodal reasoning alone. Future work may explore hybrid frameworks that combines domain-specific extractors with structured reasoning modules, while preserving the anatomical and morphological integrity required for clinically meaningful ECG interpretation.

## Abbreviations

ECG: Electrocardiogram; CNN: Convolutional Neural Network; LLM: Large Language Model; CBM: Concept Bottleneck Model; CBAM: Convolutional Block Attention Module; MHSA: Multi-Head Self Attention; MLP: Multi-Layer Perceptron; EMA: Exponential Moving Average; CI: Confidence Interval; SD: Standard Deviation; CDF: Cumulative Distribution Function; ROC-AUC: Receiver Operating Characteristic – Area Under the Curve; PR-AUC: Precision-Recall – Area Under the Curve.

## Availability of data and materials

The primary datasets are publicly available, and the dataset-associated publications are: [29, 30]. The generated data used and analysed during this study are available from the corresponding author on reasonable request.

## Ethics approval and consent to participate

Not applicable for this project.

## Funding

Funding from the Australian Research Council (FL240100217) is gratefully acknowledged.

## Competing Interests

The authors declare that they have no competing interests.